%% file: main.tex
\documentclass{article}

 \usepackage[preprint]{neurips_2023}

\usepackage{amsmath}
\usepackage{multirow}
\usepackage{float}
\usepackage{appendix}
\usepackage{graphicx}
\usepackage{subfigure}
\usepackage{amsfonts, amssymb}
\usepackage{booktabs} 
\usepackage{algorithm}
\usepackage{algpseudocode}

\newtheorem{definition}{Definition}
\newtheorem{prop}{Proposition}

\usepackage[utf8]{inputenc} 
\usepackage[T1]{fontenc}    
\usepackage{hyperref}       
\usepackage{url}            
\usepackage{booktabs}       
\usepackage{amsfonts}       
\usepackage{nicefrac}       
\usepackage{microtype}      
\usepackage{xcolor}         

\title{Augmented Message Passing Stein Variational Gradient Descent}

\author{%
  Jiankui Zhou\\
  School of Information Science and Technology\\
  ShanghaiTech University\\
  \texttt{zhoujk1@shanghaitech.edu.cn} \\
  \And
  Yue Qiu\thanks{Funded by NSF China under No. 12101407} \\
  College of Mathematics and Statistics \\
  Chongqing University \\
  \texttt{qiuyue@cqu.edu.cn} \\
}

\begin{document}

\maketitle

\begin{abstract}
\input{abstract}
\end{abstract}

\input{introduction}
\input{prelim}
\input{analysis}

\input{aump}

\input{experiments}
\input{conclusion}

\bibliographystyle{plainnat}
\bibliography{main}

\clearpage
\input{appendix}

\end{document}

%% file: abstract.tex
Stein Variational Gradient Descent (SVGD) is a popular particle-based method for Bayesian inference. However, its convergence suffers from the variance collapse, which reduces the accuracy and diversity of the estimation. In this paper, we study the isotropy property of finite particles during the convergence process and show that SVGD of finite particles cannot spread across the entire sample space. Instead, all particles tend to cluster around the particle center within a certain range and we provide an analytical bound for this cluster. To further improve the effectiveness of SVGD for high-dimensional problems, we propose the Augmented Message Passing SVGD (AUMP-SVGD) method, which is a two-stage optimization procedure that does not require sparsity of the target distribution, unlike the MP-SVGD method. Our algorithm achieves satisfactory accuracy and overcomes the variance collapse problem in various benchmark problems.

%% file: introduction.tex
\section{Introduction}
Stein variational gradient descent (SVGD) proposed by~\citet{liu2016stein} is a non-parametric inference method. In order to approximate an intractable but distinguishable target distribution, it constructs a set of particles, which can be initialized from any initial distribution. These particles move in the Reproducing Kernel Hilbert Space (RKHS) determined by the kernel function. SVGD drives these particles in the direction that the KL divergence of the two distribution decreases most rapidly. SVGD  is more efficient than traditional Markov chain Monte Carlo (MCMC) method due to the fact that the particles in SVGD reach the target distribution precisely through the dynamic process. These advantages make SVGD appealing and it has attracted lots of research interest~(\citet{liu2022grassmann,ba2021understanding, zhuo2018message, salim2022convergence, YanZhou21}).

Although SVGD succeeds in many applications~(\citet{liu2017stein, yoon2018bayesian, YanZhou21a}), it lacks the necessary theoretical support in terms of convergence under the condition of limited particles. The convergence of SVGD is guaranteed only in the mean field assumption and when the number of particles is infinite, i.e., particles converge to the true distribution when the number of particles is infinite~(\citet{liu2017stein, salim2022convergence}). However, the convergence of SVGD with finite particles is still an open problem. Furthermore, it has been observed that as the dimension of the problem increases, the variance estimated by SVGD may be much smaller than that of the target distribution. This phenomenon is known as the variance collapse, which limits the applicability of SVGD due to the following facts. First, underestimated variance leads to a failed explanation of the uncertainty of the model. Second, the Bayesian inference is usually high-dimensional in practice, but SVGD is not applicable in some scenarios due to this high-dimensional curse. For example, training Bayesian Neural Network (BNNs)~(\citet{mackay1992practical}) requires inferring a huge amount of network weight post-test distributions whose dimensions are in millions~(\citet{krizhevsky2017imagenet}). Recently, structural prediction of proteins with long structures requires inferences on the position of each atom, which results in a high-dimensional problem~(\citet{wu2022protein}).

The first contributions of this paper is that we show particles of SVGD under convergence do not spread across the whole probability space but in a finite range. We give an analytic bound of this clustered space. This bounded distribution of particles is an indication of the curse of high dimension. In addition, we provide an estimate of the error between the covariance of finite particles and the true covariance.

There have been many efforts to make SVGD applicable for high-dimension problems. According to~\citet{zhuo2018message}, the size of the repulsive force of particles is inversely proportional to the dimension of the problem. Reducing the dimension of the problem is the key to address the variance collapse, such methods include combining the Grassmann manifold and matrix decompositions to reduce the dimension of the target distribution~(\citet{chen2020projected,liu2022grassmann}). Another approach to resolve this problem is to find the Markov blanket for each dimension of the target distribution, therefore the global kernel function could be replaced by a local kernel function. Under such scenarios, the efficiency of SVGD is improved and such method is called message passing SVGD (MP-SVGD)~(\citet{zhuo2018message, wang2018stein}). However, MP-SVGD needs to know the probability graph model structure in advance and is efficient only when the graph is sparse. Moreover, identifying the Markov blanket for high-dimension problems is challenging.

The second contribution of this paper is that we further overcome the shortcomings of MP-SVGD and extend MP-SVGD to high-dimension problems. Combined with the important results of our variance analysis, we propose the so-called Augmented MP-SVGD (AUMP-SVGD). AUMP-SVGD decomposes the problem dimension into three parts by the KL divergence factorization. Different from MP-SVGD, AUMP-SVGD adopts a two-stage update procedure to solve the dependence on sparse probabilistic graphical models. Therefore, it overcomes the shortcomings of the variance collapse and does not require prior knowledge of the graph structure. We show the superiority of AUMP-SVGD theoretically and experimentally to state-of-the-art algorithms.

%% file: prelim.tex
\section{Preliminaries}


\subsection{SVGD}
SVGD approximates an intractable unknown target distribution $p(\mathbf{x})$ where $\mathbf{x} = [x_{1},...,x_{D}]^{T}\in\mathcal{X}\subseteq \mathbb{R}^{D}$ with the best $q^{*}(\mathbf{x})$ by minimizing the Kullback-Leibler (KL) divergence~(\citet{liu2016stein}). Here, $D$ is the dimension of the target distribution and 
\begin{equation*}
q^{*}(\mathbf{x})=\mathop{\arg\min}\limits_{q(\mathbf{x})\in\mathcal{Q}}\mathrm{KL}(q(\mathbf{x}) \| p(\mathbf{x})).   
\end{equation*}
SVGD takes a group of particles $\left\{ \mathbf{x} \right\}^{m}_{i=0}$ from the initial distribution $q_{0}(\mathbf{x})$, and after a series of smooth transforms, these particles finally converge to the target distribution $p(\mathbf{x})$. Each smooth transformation can be expressed by $\mathbf{T(x)} = \mathbf{x} + \epsilon \Phi(\mathbf{x})$, where $\epsilon$ is the step size and $\Phi: \mathcal{X} \to \mathbb{R}^D$ is the transformation direction. Here, $\mathcal{X}$ is the collection of the particles. Let $\mathcal{H}$ denote the set of positive definite kernels $k\left(\cdot , \cdot\right)$ in the reproducing kernel Hilbert space (RKHS) and denote $\mathcal{H}^{D}=\mathcal{H} \times \cdots \times \mathcal{H}$, where $\times$ is the Cartesian product. The steepest descent direction is obtained by minimizing the KL divergence, 
\begin{equation*}
\begin{small}
\min _{\|\phi\|_{\mathcal{H}^{D} \leq 1}} \nabla_{\epsilon} \mathrm{KL}\left(q_{[T]} \| p\right)|_{\epsilon=0}=-\max _{\|\phi\|_{\mathcal{H}^{D}} \leq 1} E_{\mathbf{x} \sim q}\left[\mathcal{A}_{p} \phi(\mathbf{x})\right], 
\end{small}
\end{equation*}
where $q_{[T]}$ means particles take the distribution $q$ after $\mathbf{T(x)}$ mapping, $\mathcal{A}_{p}$ is the Stein operator given by $\mathcal{A}_{p}\phi(\mathbf{x})=\nabla_{\mathbf{x}} \log p(\mathbf{x}) \phi(\mathbf{x})^{T}+\nabla_{\mathbf{x}}\phi(\mathbf{x})$. SVGD updates the particles $\left\{\mathbf{x} \right\}_{i=1}^{m}$ drawn from the initial distribution $q_{0}(\mathbf{x})$ by
\begin{equation*}
\begin{small}
\mathbf{x}_{n}^{\ell+1}=\mathbf{x}_{n}^{\ell}+\epsilon_{l} \hat{\phi}_{\ell}^{*}\left(\mathbf{x}_{n}^{\ell}\right), \quad n=1, \ldots, m, \ell=0,1, \ldots  
\end{small}
\end{equation*}
The steepest descent direction is given by
\begin{equation}
\begin{small}
\hat{\phi}_{\ell}^{*}\left(\mathbf{x}_{n}^{\ell}\right)=\frac{1}{m} \sum_{i=1}^{m} k\left(\mathbf{x}_{i}^{\ell}, \mathbf{x}_{n}^{\ell}\right)\nabla_{\mathbf{x}_{i}^{\ell}} \log p\left(\mathbf{x}_{i}^{\ell}\right) +\nabla_{\mathbf{x}_{i}^{\ell}} k\left(\mathbf{x}_{i}^{\ell}, \mathbf{x}_{n}^{\ell}\right). \label{1}
\end{small}
\end{equation}
The kernel function can be chosen as the RBF kernel $k(\mathbf{x}, \mathbf{y})=\exp(-\|\mathbf{x} - \mathbf{y}\|_{2}^{2}/(2h))$~(\citet{liu2016stein}) or the IMQ kernel $k(\mathbf{x}, \mathbf{y}) = 1/\sqrt{1 + \|\mathbf{x} - \mathbf{y}\|_{2}^{2}/(2h)}$~(\citet{gorham2017measuring}). Equation~\eqref{1} can be divided into two parts: the driving force term $\left[k\left(\boldsymbol{x}^{\prime}, \boldsymbol{x}\right) \nabla_{\boldsymbol{x}^{\prime}} \log p\left(\boldsymbol{x}^{\prime}\right)\right]$ and the repulsive force term $\nabla_{\boldsymbol{x}^{\prime}} k\left(\boldsymbol{x}^{\prime}, \boldsymbol{x}\right)$. It has been demonstrated that SVGD falls under the high-dimension curse~(\citet{zhuo2018message,liu2017stein}). Related research shows that there exists a negative correlation between the problem dimension and the repulsive force of SVGD~(\citet{ba2021understanding}). The influence of repulsive force is mainly related to the dimension of the target distribution.

\textbf{MP-SVGD}. The Message Passing SVGD (MP-SVGD)~(\citet{zhuo2018message,wang2018stein}) is proposed to reduce the dimension of the target distribution by identifying the Markov blanket for problems with a known graph structure. For dimension index $d$, its Markov blanket $\Gamma_{d}=\cup\{F: d \in F \} \backslash\{d\}$ contains neighborhood nodes of  $d$ such that $p\left(x_{d} \mid \mathbf{x}_{\neg d}\right)=p\left(x_{d} \mid \mathbf{x}_{\Gamma_{d}}\right)$, where $F = \{1,..., D\}$ and $\neg d = \{1,..., D\} \backslash\{d\}$. However, it is necessary to rely on the sparse correlation of variables of the target distribution in order to obtain good results.

\subsection{Mixing for random variables}
Since SVGD forms a dynamic system where particles interact with each other, one can no longer treat the converged particles as independent and identically distributed. Therefore, we need to use the mathematical tool ``mixing'', cf.~\citet{bradley2005basic} for more information.

\textbf{Mixing}. Let $\left\{\mathbf{x}_{m}, m \geq 1\right\}$ be a sequence of random variables over some probability space $(\Omega, F, P)$, where $\sigma(\mathbf{x}_{i}, i \leq j)$ denotes the  $\sigma$-algebra. For any two  $\sigma$-fields $\sigma\left(\mathbf{x}_{i}, i \leq j\right),\ \sigma\left(\mathbf{x}_{i}, i \geq j+k\right)$ with any $k \geq 1$, let
\begin{equation}
\begin{small}
\beta_{0}=1 ,  \beta_{k}=\sup _{j \geq 1} \beta \left(\sigma\left(\mathbf{x}_{i}, i \leq j\right), \sigma \left(\mathbf{x}_{i}, i \geq j+k \right) \right),  \label{2}
\end{small}
\end{equation}
where
\begin{equation*}
\begin{small}
\beta(\mathcal{A}, \mathcal{B})=\frac{1}{2} \sup\{\sum_{i \in I} \sum_{j \in J} \left|\mathbb{P}\left(A_{i} \cap B_{j}\right)-\mathbb{P}\left(A_{i}\right) \mathbb{P}\left(B_{j}\right)\right|\} 
\end{small}
\end{equation*}
is the maximum taken over all finite partitions $\left(A_{i}\right)_{i \in I}$ and $\left(B_{i}\right)_{i \in J}$ of $\Omega$. A family $\left\{\mathbf{x}_{m}, m \geq 1\right\}$ of random variables will be said to be absolutely regular (or $\beta-$mixing) if $\lim_{m \to \infty} \beta_{m}= 0$, where the coefficients of absolute regularity $\beta_{m}$ are defined in Equation~\eqref{2}~(\citet{banna2016bernstein}). These coefficients quantify the strength of dependence between the $\sigma$-algebra generated by $(\mathbf{x}_i)_{1\leq i\leq k}$ and the one generated by $(\mathbf{x}_{i})_{i\geq k+m}$ for all $k \in \mathbb{N}^{*}$. $\beta_m$ tends to zero while $m$ goes to infinity implies that the $\sigma$-algebra generated by $(\mathbf{x}_{i})_{i\geq k+m}$ is less and less dependent on $\sigma\left(\mathbf{x}_{i},i \leq k\right)$~(\citet{bradley2005basic}).

%% file: analysis.tex
\section{Covariance Analysis under \textbf{$\mathbf{\beta}$-mixing}}
\citet{ba2021understanding} analyzed the convergence of SVGD when the covariance matrix of the Gaussian distribution is an identity matrix under the assumption of near-orthogonality. On the basis of this work, we further analyze the more general form of the variance collapse. Moreover, the quantification of the variance collapse for finite particles is still an open problem to the best of our knowledge. \citet{chewi20} shows that SVGD can be viewed as a kernelized Wasserstein gradient flow of the chi-squared divergence, which makes us believe that extending our analysis in this section to other MCMC methods such as Wasserstein gradient flow or normalizing flow is possible.

\subsection{Assumptions}
\textbf{A1 (Fixed points)}. Although the convergence of SVGD under finite particles is still an open problem, it can be found in many experimental studies that all particles converge to fixed points~(\citet{ba2021understanding}). In this paper, we also assume that for SVGD with finite number of particles, these particles eventually converge to the target distribution and approach fixed points. 

\textbf{A2 ($m$-dependent of particles)} 
\begin{definition}(\citet{hoeffding1948central})
If for some function $f(n)$, the inequality $s-r\geq f(n) \geq 0$ implies that the tow sets $(x_{1},...,x_{r})$ and $(x_{s},..., x_{m})$ are independent, then the sequence $\left\{x_{i}\right\}_{i=1}^{m}$ is said to $f(n)$-dependent.  
\end{definition}

We assume that the fixed points of SVGD satisfy the $m$-dependent assumption. In~\citet{ba2021understanding}, only weak correlation between SVGD particles is reported. We perform numerical verification in Appendix B and leave the rigorous proof for our future work. Under Assumptions {\bf A1}-{\bf A2} and the zero-mean of the target distribution, we analyze the variance collapse of SVGD quantitatively in the following part.

\subsection{Concentration of particles}
We first give the upper bound of the concentrated particles. The main tool used here is the Jensen gap~(\citet{gao2017bounds}).

\begin{prop}
Let Assumption {\bf A1} hold, for mean zero Gaussian target and Gaussian RBF kernel $k(\mathbf{x}_i, \mathbf{x}_j) = \exp^{-\frac{\|\mathbf{x}_i- \mathbf{x}_j\|_{2}^{2}}{2h}}$, we have
\begin{equation*}
\begin{small}
\|\mathbf{x}_{i}\|_{2} \leq (2M/c + 1)\left(\mathbb{E}\|\mathbf{x}\|_{2}^{2}\right)^{1/2},  \label{7}
\end{small}
\end{equation*}
where $M=\sup_{\mathbf{x} \neq 0} \frac{ \left| k(\mathbf{x}_i, \mathbf{x}) - k(\mathbf{x}_i, 0) \right|} {2\|\mathbf{x}\|_2}$, $c=\frac{2}{h}e^{\frac{-c_0 tr(\Sigma_m)}{h}}$,  $1<c_0\leq m$ is a positive constant, and $\Sigma_m$ is the empirical covariance matrix of the particles.
\end{prop}

We leave this proof in Appendix A. For most of the sampling-based inference methods, such as MCMC, VI, et al., samples spread across the whole sample space but with extremely small probability for some samples. However, Proposition 1 shows that SVGD with finite particles are confined to a certain range, although this range may expand as the number of particles increases. With the RBF kernel, this upper bound is related to the trace of the covariance of the target distribution. Under the IMQ kernel $k(\mathbf{x}, \mathbf{y}) = 1/\sqrt{1 + \|\mathbf{x} -\mathbf{y}\|_{2}^{2}/(2h)}$ conditions, this range will be further reduced~(\citet{gorham2017measuring}).

\subsection{Covariance estimation}

For independent and identically distributed (i.i.d.) samples, the variance can be estimated using the Bernstein inequality or the Lieb inequality. However, these random matrix results typically require the particles to be i.i.d, which is no longer satisfied by SVGD due to its interacting update. To analyze the variance of particles from SVGD, we assume these particles are $m$-dependent. For a sequence of random variables, the $m$-dependent assumption implies that they satisfy the $\beta$-mixing condition~(\citet{bradley2005basic}). We obtain the following results based on the Bernstein inequality of dependent random matrices~(\citet{banna2016bernstein}).

\begin{prop}\label{prop_2}
Let Assumption {\bf A1}-{\bf A2} hold, for SVGD with the Gaussian RBF kernel $k(\mathbf{x}_i, \mathbf{x}_j) = \exp^{-\frac{\|\mathbf{x}_i- \mathbf{x}_j\|_{2}^{2}}{2h}}$,  denote $\mathbb{X}_{i}=\mathbf{x}_i\mathbf{x}_i^T - \Sigma$ where $\Sigma
$ is the covariance matrix of the target distribution. There exists $\alpha \geq 0$ such that for any integer $k \geq 1$, the following inequality holds,
\begin{equation*}
\begin{small}
\beta_{k}=\sup _{j \geq 1} \beta\left(\sigma\left(\mathbf{x}_{i}, i \leq j\right), \sigma\left(\mathbf{x}_{i}, i \geq j+k\right)\right) \leq \mathrm{e}^{-\alpha (k-1)}. 
\end{small}
\end{equation*}
Denote the covariance matrix of these $m$ particles by $\Sigma_m$, we have
\begin{equation*}
\mathbb{E} \| \Sigma_m - \Sigma\| \leq  30v \sqrt{\frac{ \log D }{m}}+\frac{2K^2 tr(\Sigma)}{m}(4 \alpha ^{-1/2}\sqrt{\log D}+\gamma(\alpha , m) \log D),  \label{covv}
\end{equation*}
where $K = (2M/c + 1)$ and $M$, $c$ are identical to that in Proposition 1. Here, $\alpha$ measures the correlation between particles, and $D$ is the dimension of the target distribution. Moreover,
\begin{equation*}
\begin{small}
v^{2}=\sup _{N \subseteq\{1, \ldots, m\}} \frac{1}{|N|} \lambda_{\max }\left(\mathbb{E}(\sum_{i \in N} \mathbb{X}_{i})^{2}\right).  \label{10}
\end{small}
\end{equation*}
Here, $\lambda_{\max}\left(\mathbb{X}\right)$ represents the eigenvalue of  $\mathbb{X}$ with the maximum magnitude, $|N|$ is the cardinality of the set $N$, and
\begin{equation*}
\begin{small}
\gamma(\alpha,\ m)=\frac{\log m}{\log 2} \max \left(2, \frac{32 \log m}{\alpha  \log 2}\right).
\end{small}
\end{equation*}
\end{prop}
Proposition 2 shows that the main factors that affect the upper bound of the variance error include the inter-particle correlation, the number of particles, the dimension of the target distribution, and the trace of its covariance matrix. According to~\citet{ba2021understanding}, it can be considered that $tr(\Sigma_m) \leq tr(\Sigma)$ should be true for SVGD, therefore the constant $c$ in Proposition 1 can be replaced by 
$\frac{2}{h} e^{\frac{-c_0 tr(\Sigma)}{h}}$.

%% file: aump.tex
\section{Augmented MP-SVGD}
Here, we propose the so-called Augmented Message Passing (AUMP-SVGD) to overcome the covariance underestimation of SVGD. Compared with MP-SVGD, AUMP-SVGD requires neither a known graph structure nor the sparsity structure of the target distribution.

\subsection{MP-SVGD}
The update direction $\Delta$ of SVGD is given by
\begin{equation*}
\begin{small}
\Delta^{\mathrm{SVGD}}(\boldsymbol{x})=\mathbb{E}_{\boldsymbol{x}^{\prime} \sim q}\underbrace{k\left(\boldsymbol{x}^{\prime}, \boldsymbol{x}\right) \nabla_{\boldsymbol{x}^{\prime}} \log p\left(\boldsymbol{x}^{\prime}\right)}_{\text{driving force
}}+\underbrace{\nabla_{\boldsymbol{x}^{\prime}}k\left(\boldsymbol{x}^{\prime},\boldsymbol{x}\right)}_{\text{repulsive force}}
\end{small}
\end{equation*}
The log derivative term in the update rule $\mathbb{E}_{\boldsymbol{x}^{\prime} \sim q}\left[k(\boldsymbol{x}^{\prime}, \boldsymbol{x}) \nabla_{\boldsymbol{x}^{\prime}} \log p(\boldsymbol{x}^{\prime})\right]$ corresponds to the driving force that guides particles toward the high-likelihood region. The (Stein) score function $\nabla_{\boldsymbol{x}^{\prime}} \log p\left(\boldsymbol{x}^{\prime}\right)$ is a vector field describing the target distribution. 
$\nabla_{\boldsymbol{x}^{\prime}}k\left(\boldsymbol{x}^{\prime}, \boldsymbol{x}\right)$ provides a repulsive force to prevent particles from aggregating. However, as the dimension of the target distribution increases, this repulsive force gradually decreases~(\citet{zhuo2018message}). This causes SVGD to fall under the curse of high dimensions. How to effectively reduce the dimension has become the guiding ideology to make SVGD overcome the curse of dimensionality.

Our concern is the problem with the continuous graphical model, i.e., the target distribution that satisfies the following form $p(\mathbf{x}) \propto \exp \left[\sum_{s \in \mathcal{S}} \psi\left(\mathbf{x}_{s}\right)\right]$ where $\mathcal{S}$ is the family of index sets $s \subseteq\{1, \ldots, D\}$ that specifies the Markov structure. For any index $i$, its Markov blanket is represented by $\mathcal{N}_{i}:=\cup\{s: s \subset \mathcal{S}, i \in s \} \backslash\{i\}$. According to~\citet{wang2018stein}, one can transform the global kernel function into a local kernel function and this local kernel function is just related to $\mathcal{C}_{i}:=\mathcal{N}_{i} \cup\{i\}$ for any $i$. Under this transformation, the dimension is reduced from $D$ to $|\mathcal{C}_{i}|$, where $|\mathcal{C}_{i} |$ is the size of the set $\mathcal{C}_{i}$. Then, SVGD becomes
\begin{subequations}
\begin{align}
&\mathbf{x}_{i}^{\ell}\leftarrow \mathbf{x}_{i}^{\ell}+\epsilon \phi^{*}\left(\mathbf{x}_{i}^{\ell}\right), \quad \forall i \in \{1,...,D\}, \ell \in \{1,...,m\},  \label{14}\\
&\phi^{*}(\mathbf{x}_{i}^{\ell}) :=\frac{1}{m} \sum_{\ell=1}^{m} s\left(\mathbf{x}_{i}^{\ell}\right) k\left(\mathbf{x}_{i}^{\ell}, \mathbf{x}_{\mathcal{C}_{i}}\right)+\partial_{\mathbf{x}_{i}^{\ell}} k_{i}\left(\mathbf{x}_{i}^{\ell}, \mathbf{x}_{\mathcal{C}_{i}}\right),   \label{15}
\end{align}
\end{subequations}
where $s(x_{i}^{\ell})=\nabla_{x_{i}^{\ell}} \log p(x_i^{\ell})$. Such method is known as the message passing SVGD (MP-SVGD)~(\citet{zhuo2018message, liu2017stein}). However, MP-SVGD needs to know the graph structure of the target distribution in advance to determine the Markov blanket and the sparsity of this graph is needed in order to achieve dimension reduction.

\subsection{Augmented MP-SVGD}
Inspired by MP-SVGD, we propose the augmented MP-SVGD which is suitable for more complex graph structures. We keep the previous symbols but redefine them for clarity. We assume $p(\mathbf{x})$ can be factorized as $p(\mathbf{x}) \propto \prod_{F \in \mathcal{F}} \psi_{F}\left(\mathbf{x}_{F}\right)$ where $F \subset\{1, \ldots, D\}$ is the index set. Partition $\Gamma_{d}\subset\{1, \ldots, D\} \backslash\{d\}$ and $\mathbf{S_{d}} \subset\{1, \ldots, D\} \backslash\{d\}$ such that $\Gamma_{d} \cap \mathbf{S_{d}} = \emptyset$ and $\Gamma_{d} \cup \mathbf{S_{d}} =\{1, \ldots, D\}\backslash \{d\}$. Let  $\mathbf{x}_\mathbf{S_{d}}= [x_i,\ ...\ ,\ x_j],\ i,...,j \in \mathbf{S_{d}}$. Similarly, $\mathbf{x}_{\Gamma_{d}}= [x_i,\ ...\ ,\ x_j],\  i,...,j \in \Gamma_{d}$. 

\begin{figure}[h]
\centering
\centerline{\includegraphics[width=0.65\textwidth]{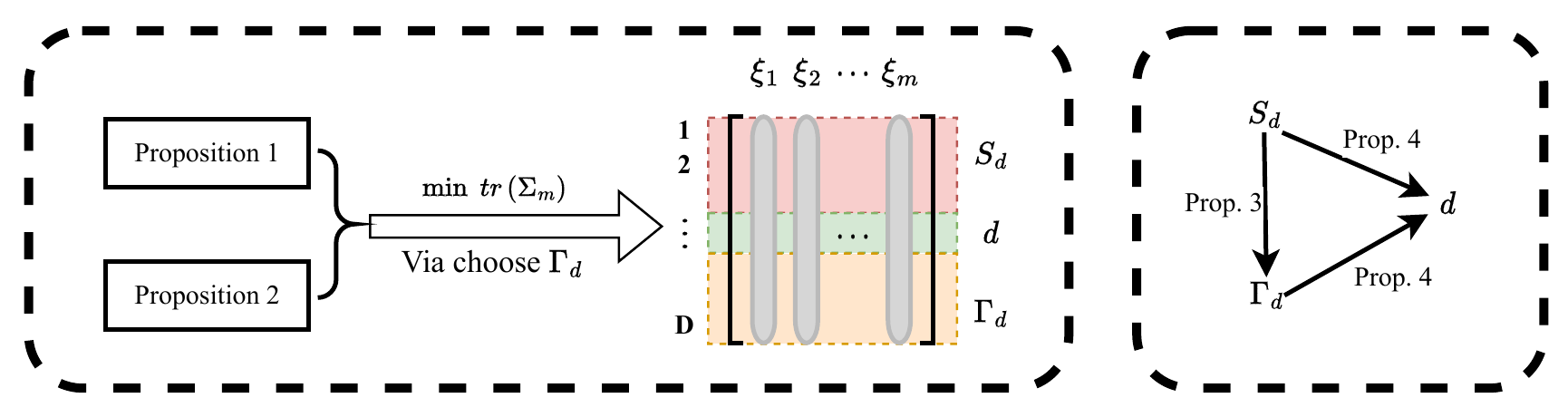}}
\caption{Probabilistic Graphical Model}\label{graph-model}
\end{figure}

Consider the probability graph model illustrated by Figure~\ref{graph-model}, $p(\mathbf{x})$ can be represented as $p(\mathbf{x})=p(x_{d}|\mathbf{x}_{\Gamma_{d}},\mathbf{x}_{\mathbf{S_{d}}})p(\mathbf{x}_{\Gamma_{d}}|\mathbf{x}_{\mathbf{S_{d}}})p(\mathbf{x}_{\mathbf{S_{d}}})$. Our method relies on the key observation of~\citet{zhuo2018message} that
\begin{equation*}\label{15.deom}
\begin{small}
\mathrm{KL}(q \| p)=\underbrace{\mathrm{KL} \left(q\left(\mathbf{x}_{\Gamma_{d}}, \mathbf{x}_{\mathbf{S_{d}}}\right) \|p\left(\mathbf{x}_{\Gamma_{d}},\mathbf{x}_{\mathbf{S_{d}}}\right)\right)}_{\text{prKL}(q, p, d)} +\underbrace{\mathrm{KL}\left(q\left(x_{d}\mid\mathbf{x}_{\neg d}\right)q\left(\mathbf{x}_{\neg d}\right)\|p\left(x_{d}\mid\mathbf{x}_{\neg d}\right)p\left(\mathbf{x}_{\neg d}\right) \right)}_{\text{seKL}(q, p, d)}. 
\end{small}
\end{equation*}   

To minimize $\mathrm{KL}(q\|p)$, we adopt a two-stage procedure so that $\text{prKL}(q, p, d)$ and $\text{seKL}(q, p, d)$ are optimized alternatively. At the first stage, $\text{prKL}(q, p, d)$ is further decomposed into 
\begin{equation}\label{16}
\text{prKL}(q, p, d) = \mathrm{KL} \left(q\left(x_{\mathbf{S_{d}}}\right) \|p\left(x_{\mathbf{S_{d}}}\right)\right) + \mathrm{KL}\left(q\left(\mathbf{x}_{\Gamma_{d}} \mid x_{\mathbf{S_{d}}}\right)q\left(x_{\mathbf{S_{d}}}\right) \| p\left(\mathbf{x}_{\Gamma_{d}}\mid x_{\mathbf{S_{d}}}\right) p\left(x_{\mathbf{S_{d}}}\right)\right).  
\end{equation}
We can fix $\mathbf{x}_{\mathbf{S_{d}}}$ to apply the local kernel function to the second part of Equation~\eqref{16} to minimize $\text{prKL}(q, p, d)$, 
\begin{equation*}
\begin{small}
\begin{aligned}
\mathbf{x}_{\Gamma_{d}} = \mathop{\arg\min}\limits_{q\left(\mathbf{x}_{\Gamma_{d}}\mid \mathbf{x}_{\mathbf{S_{d}}}\right)}\mathrm{KL}\left(q\left(\mathbf{x}_{\Gamma_{d}}\mid \mathbf{x}_{\mathbf{S_{d}}}\right) q\left(\mathbf{x}_{\mathbf{S_{d}}}\right) \| p\left(\mathbf{x}_{\Gamma_{d}} \mid \mathbf{x}_{\mathbf{S_{d}}}\right) p\left(\mathbf{x}_{\mathbf{S_{d}}}\right)\right). \label{q1}
\end{aligned} 
\end{small}
\end{equation*}
This optimization procedure is given by Proposition~\ref{prop_3} and we leave the proof in the appendix.

\begin{prop}\label{prop_3}
Let $T(\mathbf{x})=\left[x_{1},...,T_{\Gamma_{d}}\left(\mathbf{x}_{\Gamma_{d}}\right),..., x_{D}\right]^{T}$ where $T_{\Gamma_{d}}:\ \mathbf{x}_{\Gamma_{d}} \rightarrow \mathbf{x}_{\Gamma_{d}}+\epsilon \phi_{\Gamma_{d}}\left(\mathbf{x}_{\neg d}\right)$ and $\phi_{\Gamma_{d}} \in \mathcal{H}_{\Gamma_{d}}$. Here $\mathcal{H}_{\Gamma_{d}}$ is the space that defines the local kernel function $k_{\Gamma_{d}}:\ \mathcal{X}_{\neg d} \times \mathcal{X}_{\neg d} \rightarrow \mathbb{R}$. The optimal solution of the optimization problem
\begin{equation*}\label{argmin}
\begin{small}
\begin{aligned}
\mathop{\min}\limits_{\left\|\phi_{\Gamma_{d}}\right\| \leq 1} \left. \nabla_{\epsilon} \mathrm{KL}\left(q_{[T]}\left(\mathbf{x}_{\Gamma_{d}},x_{\mathbf{S_{d}}}\right) \|p\left(\mathbf{x}_{\Gamma_{d}},x_{\mathbf{S_{d}}}\right)\right) \right|_{\epsilon=0},    
\end{aligned}
\end{small}
\end{equation*}
is given by $\phi_{\Gamma_{d}}^{*} /\left\|\phi_{\Gamma_{d}}^{*}\right\|_{\mathcal{H}_{\Gamma_{d}}}$ where
\begin{equation*}
\begin{small}
\phi_{\Gamma_{d}}^{*}\left(\mathbf{x}_{\neg d}\right)= \mathbb{E}_{\mathbf{y}_{\neg d} \sim q} [k_{\Gamma_{d}}\left(\mathbf{x}_{\neg d}, \mathbf{y}_{\neg d}\right) \nabla_{\mathbf{y}_{\Gamma_{d}}} \log p\left(\mathbf{y}_{\Gamma_{d}} \mid \mathbf{y}_{\neg d}\right)+\nabla_{\mathbf{y}_{\Gamma_{d}}} k_{\Gamma_{d}}\left(\mathbf{x}_{\neg d}, \mathbf{y}_{\neg d}\right)]. 
\end{small}
\end{equation*}
\end{prop}

At the second stage, $\mathbf{x}_{\Gamma_{d}}$ and $\mathbf{x}_{\mathbf{S_{d}}}$ are fixed while only $\mathbf{x}_{d}$ is updated. We can further decompose $\mathrm{seKL}(q,p,d)$ into three parts via the convexity of the KL divergence,
\begin{equation*}\label{dec}
\begin{small}
0 \leq \mathrm{seKL}(q,p,d)\leq \mathrm{KL}\left[\frac{q(x_{d} \mid \mathbf{x}_{\Gamma_{d}})} {q(\mathbf{x}_{\neg d})} \| \frac{p(x_{d} \mid \mathbf{x}_{\Gamma_{d}})} {p(\mathbf{x}_{\neg d})}\right]+ \mathrm{KL}\left[\frac{q(x_{d} \mid \mathbf{x}_{\mathbf{S_{d}}} )} {q(\mathbf{x}_{\neg d})}  \|  \frac{p(x_{d} \mid \mathbf{x}_{\mathbf{S_{d}}})} {p(\mathbf{x}_{\neg d})} \right] + C, 
\end{small}
\end{equation*}
where $C =\mathrm{KL}\left[\frac{1} {q(\mathbf{x}_{\neg d})} \| \frac{1} {p(\mathbf{x}_{\neg d})}\right]$ is a positive constant due to the fact that $\mathbf{x}_{\neg d} =\mathbf{x}_{\Gamma_{d}}\cup  \mathbf{x}_{\mathbf{S_{d}}}$ is fixed. Therefore, 
\begin{equation*}
\begin{small}
x_{d} = 
\mathop{\arg\min}\limits_{q(x_{d} \mid \mathbf{x}_{\Gamma_{d}})),p(x_{d} \mid \mathbf{x}_{\mathbf{S_{d}}})} \mathrm{KL}\left[\frac{q(x_{d} \mid \mathbf{x}_{\Gamma_{d}})} {q(\mathbf{x}_{\neg d})} \| \frac{p(x_{d} \mid \mathbf{x}_{\Gamma_{d}})} {p(\mathbf{x}_{\neg d})}\right]+ 
\mathrm{KL} \left[\frac{q(x_{d} \mid \mathbf{x}_{\mathbf{S_{d}}} )} {q(\mathbf{x}_{\neg d})}  \| \frac{p(x_{d} \mid \mathbf{x}_{\mathbf{S_{d}}})} {p(\mathbf{x}_{\neg d})} \right].  \label{q2}
\end{small}
\end{equation*}
The optimization procedure is described by Proposition~\ref{prop_4} and we also leave the proof in the appendix. 

\begin{prop}\label{prop_4}
Let $T(\mathbf{x}) = \left[x_{1}, \ldots, T_{d}\left(x_{d}\right), \ldots, x_{D}\right]^{T}$ where $T_{d}: x_{d} \rightarrow x_{d}+\epsilon \phi_{d}\left(\mathbf{x}_{C_{d}}\right)$ and $\phi_{d} \in \mathcal{H}_{d}$. Here $\mathcal{H}_{d}$ is the space that defines the local kernel $k_{d}: \mathcal{X}_{S_{d}} \times \mathcal{X}_{S_{d}} \rightarrow \mathbb{R}$ and $C_{d} = S_{d} \cup \{d\}$ or $C_{d} = \Gamma_{d} \cup \{d\}$. The optimal solution of the following optimization problem
\begin{equation*}
\begin{small}
\mathop{\min}\limits_{\left\|\phi_{d}\right\| \leq 1} \left.  
\nabla_{\epsilon} \mathrm{KL} \left[\frac{q_{[T]}(x_{d} \mid \mathbf{x}_{\mathbf{C_{d}}} )} {q(\mathbf{x}_{\neg d})}  \| \frac{p(x_{d} \mid \mathbf{x}_{\mathbf{C_{d}}})} {p(\mathbf{x}_{\neg d})} \right]   \right|_{\epsilon=0},  \label{p2argmin}    
\end{small}
\end{equation*}
is given by $\phi_{d}^{*} / \left\|\phi_{d}^{*}\right\|_{\mathcal{H}_{d}}$ where
\begin{equation*}
\begin{small}
\phi_{d}^{*}\left(\mathbf{x}_{C_{d}}\right) =  \mathbb{E}_{\mathbf{y}_{C_{d}} \sim q} \left[k_{d}\left(\mathbf{x}_{C_{d}}, \mathbf{y}_{C_{d}}\right) \nabla_{y_{d}} \log p \left(y_{d} \mid \mathbf{y}_{C_{d}} \right) +\nabla_{y_{d}} k_{d}\left(\mathbf{x}_{C_{d}}, \mathbf{y}_{C_{d}}\right)\right]. 
\end{small}
\end{equation*}
\end{prop}

For $d\in \left\{1, 2, \dots, D \right\}$, $x_{d}$ is updated through the above two-stage procedure, which reduces the dimension of the original problem from $D$ to the size of $min(|\mathbf{x}_{\mathbf{S_{d}}}|,\ |\mathbf{x}_{\Gamma_{d}}|)$. In this way, we are able to solve the problem of inferring more complex target distributions that traditional MP-SVGD failed. This comparison is illustrated in Experiment 2. Moreover, we do not need to know the real probability graph structure of the target distribution in advance. 

The key start of our AUMP-SVGD is to choose $\Gamma_{d}$ or $\mathbf{S_{d}}$. This problem can be formulated in the following form, for $m$ particles with each $x \in \mathbb{R}^D$, denote the ensemble matrix of these particles by $X$, i.e., $X \in \mathbb{R}^{m \times D}$. Let $r = |\mathbf{x}_{\Gamma_{d}}|$ and $\mathcal{M}(\mathrm{r}, m \times D)$ be the set of sub-matrices $\Gamma_x \in \mathbb{R}^{m \times r}$ of $X$. Determining the set $\Gamma_d$ corresponds to select the sub-matrix $X_r$ from the set $\mathcal{M}(\mathrm{r}, m \times D)$ to minimize $\mathbf{E} \| \Sigma_m - \Sigma_x \|$, where $\Sigma_m$ is the empirical covariance matrix of particles and $\Sigma_x$ is the empirical covariance matrix of sub-ensembles. The upper bound of $\mathbf{E} \| \Sigma_m - \Sigma_x \|$ is already given by Proposition~\ref{prop_2} and it is related to $tr(\Sigma_{x})$. Therefore, we choose the sub-matrix with the smallest $tr(\Sigma_{x})$ to ensure that $\mathbb{E} \| \Sigma_m - \Sigma_x\|$ is minimized. This corresponds to select $X_{r} \in \mathbb{R}^{m \times r}$  from  $X$  to get minimal $\operatorname{tr}\left(X_{r} X_{r}^{T}\right)$. Since $\operatorname{tr}\left(X_{r} X_{r}^{T}\right)=\operatorname{tr}\left(X_{r}^{T} X_{r}\right)$ , we just need to reorder each column of  $X$ by the 2-norm and choose $r$ columns with smallest 2-norm. The computational complexity for this part is $\mathcal{O}\left(D m^{2}+D \log D\right)$. Finally, we give the complete form of our AUMP-SVGD by Algorithm~\ref{alg:aump}.

\begin{algorithm}
{\scriptsize
    \caption{Augment Message Passing SVGD}\label{alg:aump}
    \begin{algorithmic}
    \State {\bfseries Input:} a set of initial particles $\left\{\mathbf{x}^{(i)}\right\}_{i=1}^{m}$
\For{iteration $i$}  
\For{$d \in {1,...,D}$}
\State Set $\Gamma_{d}$ and $S_{d}$ \\
\State $\mathbf{x}_{\Gamma_{d}}^{(i)} \leftarrow \mathbf{x}_{\Gamma_{d}}^{(i)}+\epsilon\hat{\phi}_{\Gamma_{d}}^{*}(\mathbf{x}_{\neg d}^{(i)})$  \\
\State  $\hat{\phi}_{\Gamma_{d}}^{*}\left(\mathbf{x}_{\neg d}\right)=\mathbb{E}_{\mathbf{y}_{\neg d} \sim \hat{q}_{M}}[k_{\Gamma_{d}}\left(\mathbf{x}_{\neg d}, \mathbf{y}_{\neg d}\right) \nabla_{y_{\Gamma_{d}}} \log p \left(\mathbf{y}_{\Gamma_d} \mid \mathbf{y}_{\neg d}\right) +\nabla_{\mathbf{y}_{\Gamma_{d}}} k_{d}\left(\mathbf{x}_{\neg d}, \mathbf{y}_{\neg d}\right)]$\\
\State update $x_d$ from $\mathbf{x}_{S_{d}}$ by $C_{d} = \mathbf{x}_{S_{d}} \cup  {x_d}$\\
\State \qquad $\mathbf{x}_{d}^{(i)} \leftarrow \mathbf{x}_{d}^{(i)}+\epsilon\hat{\phi}_{d}^{*}(\mathbf{x}_{C_{d}}^{(i)})$\\
\State \qquad$\hat{\phi}_{d}^{*}\left(\mathbf{x}_{C_{d}}\right)=\mathbb{E}_{\mathbf{y}_{C_{d}} \sim \hat{q}_{M}} \left[\nabla_{y_{d}} k_{d}\left(\mathbf{x}_{C_{d}}, \mathbf{y}_{C_{d}}\right) + k_{d}\left(\mathbf{x}_{C_{d}}, \mathbf{y}_{C_{d}}\right) \nabla_{y_{d}} \log p\left(y_{d} \mid \mathbf{y}_{S_{d}}\right)\right]$\\
\State update $x_d$ from $\mathbf{x}_{\Gamma_{d}}$ by $C_{d} = \mathbf{x}_{\Gamma_{d}} \cup  {x_d}$\\
\State  \qquad $\mathbf{x}_{d}^{(i)} \leftarrow \mathbf{x}_{d}^{(i)}+\epsilon\hat{\phi}_{d}^{*}(\mathbf{x}_{C_{d}}^{(i)})$\\
\State \qquad $\hat{\phi}_{d}^{*}\left(\mathbf{x}_{C_{d}}\right)=\mathbb{E}_{\mathbf{y}_{C_{d}} \sim \hat{q}_{M}} \left[\nabla_{y_{d}} k_{d}\left(\mathbf{x}_{C_{d}}, \mathbf{y}_{C_{d}}\right) + k_{d}\left(\mathbf{x}_{C_{d}}, \mathbf{y}_{C_{d}}\right) \nabla_{y_{d}} \log p\left(y_{d} \mid \mathbf{y}_{\Gamma_{d}}\right)\right]$\\
\EndFor
\EndFor
 \State {\bfseries Output:} a set of particles $\left\{\mathbf{x}^{(i)}\right\}_{i=1}^{m}$ as samples from $p(\mathbf{x})$
\end{algorithmic}
}
\end{algorithm}

%% file: experiments.tex
\section{Experiments}
We study the uncertainty quantification properties of AUMP-SVGD compared with other existing methods such as SVGD, MP-SVGD, projected SVGD (PSVGD)~(\citet{chen2020projected}), and Sliced Stein variational gradient descent (S-SVGD)~(\citet{gong2020sliced}) through extensive experiments. We conclude from experiments that SVGD may underestimate uncertainty, S-SVGD may overestimate it and AUMP-SVGD with a properly partitioned $\Gamma_d$ and $\mathbf{S}_d$ produces the best estimate. In almost all scenarios, AUMP-SVGD outperforms PSVGD.

\subsection{Gaussian Mixture Models}
\textbf{Multivariate Gaussian}. The first example is a $D$-dimensional multivariate Gaussian $p(\mathbf{x})=\mathcal{N}\left(0, I_{D}\right)$. 
For each method, 100 particles are initialized from $q_0(\mathbf{x})=\mathcal{N}\left(10, I_{D}\right)$. 

\begin{figure}[h]
\centering
\includegraphics[width=0.6\textwidth]{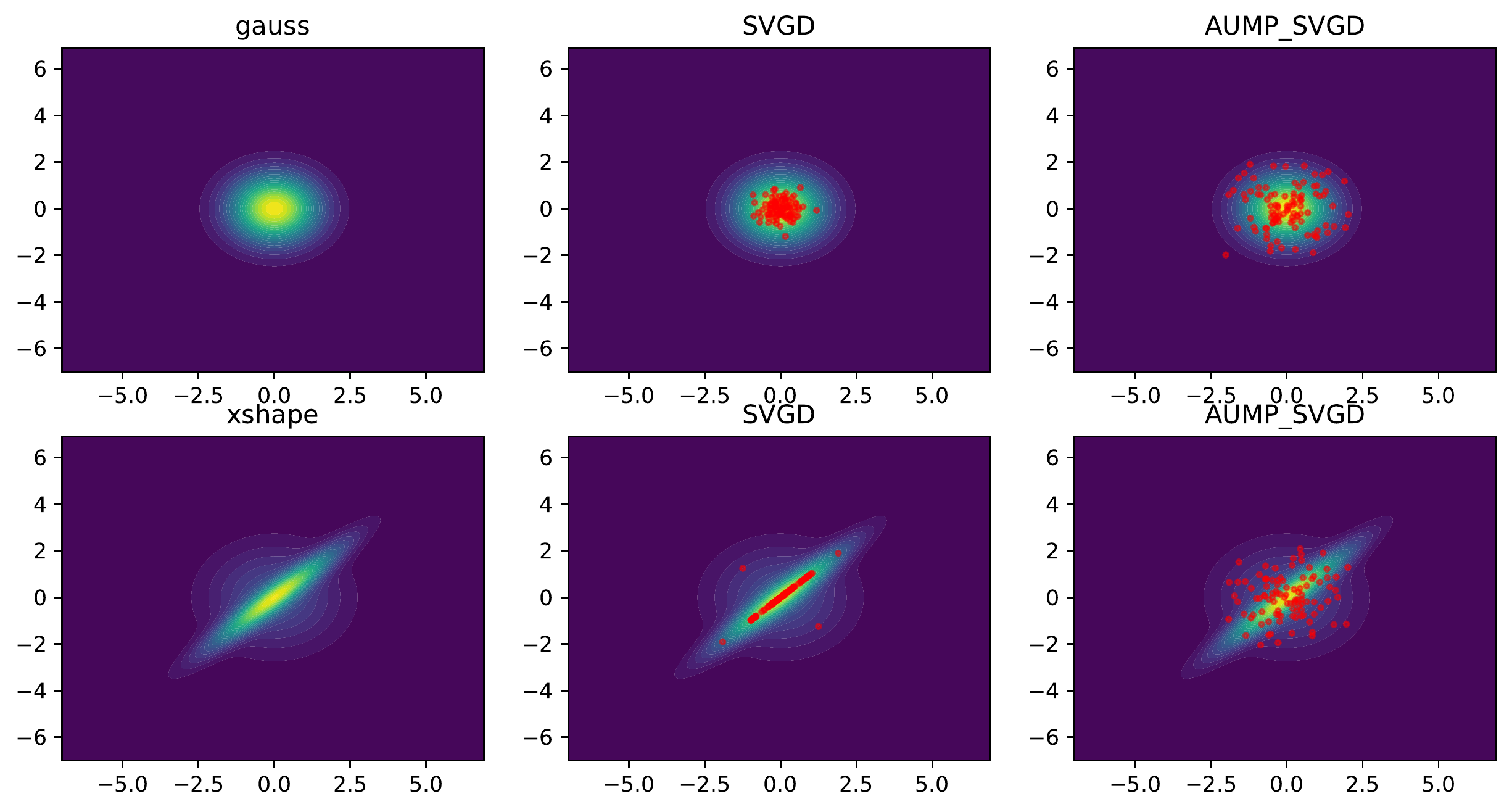}
\caption{The result of object edges and estimated densities in the first two dimensions. Red dots are particles after 2000 iterations. Both examples are 50-dimensional} 
\label{toy examples}
\end{figure}

\textbf{Spaceship Mixture}. The target in the second experiment is a $D$-dimensional mixture of two correlated Gaussian distributions $p(\mathbf{x})=0.5 \mathcal{N}\left(x;\ \mu_{1},\ \Sigma_{1}\right)+0.5 \mathcal{N}\left(x;\ \mu_{2},\ \Sigma_{2}\right)$. The mean $\mu_{1}$, $\mu_{2}$ of each Gaussian have components equal to 1 in the first two coordinates and 0 otherwise. The covariance matrix admits a correlated block diagonal structure. The mixture hence manifests as a ``spaceship'' density margin in the first two dimensions (see Figure~\ref{toy examples}). 

\vspace{-.3cm}

\begin{figure}[h]
\centering
\includegraphics[width=0.95\textwidth]{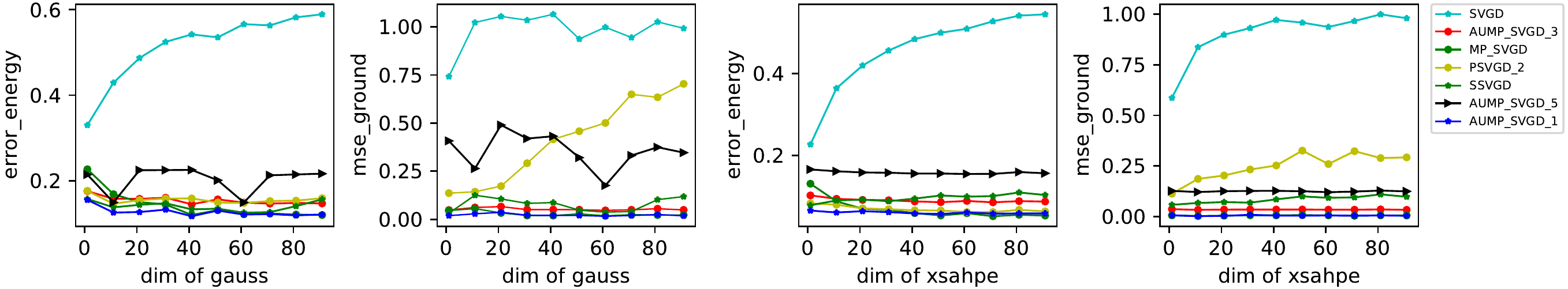}
\caption{The energy distance and MSE between SVGD samples and the real target distribution.} 
\label{toy examples vs}
\end{figure}

It can be seen from Figure~\ref{toy examples} that for the high-dimensional inference, particles from SVGD aggregate, which leads to a high-dimensional curse~(\citet{zhuo2018message,liu2016stein}). However, AUMP-SVGD can estimate the true probability distribution well in these high-dimensional situations. We calculate the energy distance and the mean-square error (MSE) $\mathbb{E}\|\Sigma_m -\Sigma\|_2$ between the samples from the inference algorithm and the real samples. The energy distance is given by $D^{2}(F, G)=2 \mathbb{E}\|X-Y\|-\mathbb{E}\left\|X-X^{\prime}\right\|-\mathbb{E}\left\|Y-Y^{\prime}\right\|$, where $F$ and $G$ are the cumulative distribution function (CDF) of $X$ and $Y$, respectively. $X^{\prime}$ and $Y^{\prime}$ denote an independent and identically distributed (i.i.d.) copy of $X$ and $Y$~(\cite{rizzo16}). 10 experiments are performed and the averaged results are given in Figure~\ref{toy examples vs}.

\vspace{-0.1cm}

Figure~\ref{toy examples vs} demonstrates the gradual expansion of the error difference of SVGD as the dimension increases. For the sparse problem, when the graph structure is already known, MP-SVGD achieves comparable outcomes with S-SVGD and PSVGD-2. In the above example, PSVGD achieves its best results when the problem dimension is reduced to 2. The AUMP-SVGD with a set size of $\Gamma_d$ of 1 or 3 outperforms other methods. In Experiment 1, we demonstrate that AUMP-SVGD yields a variance estimate equivalent to that of MP-SVGD under the simplified graph structure. In Experiment 2, as the correlation between different dimensions of the target becomes strong or the density of the graph increases, our algorithm performs better than other SVGD variants. Furthermore, our approach exhibits superior variance estimation compared with SVGD, MP-SVGD, S-SVGD, and PSVGD. In practice, the covariance matrix of the target distribution may not be sparse, making it challenging to capture this structure. As a result, the effectiveness of MP-SVGD is significantly limited. However, AUMP-SVGD can still attain stable and superior results.

\begin{figure}[h]
    \centering
    \includegraphics[width=0.6\textwidth]{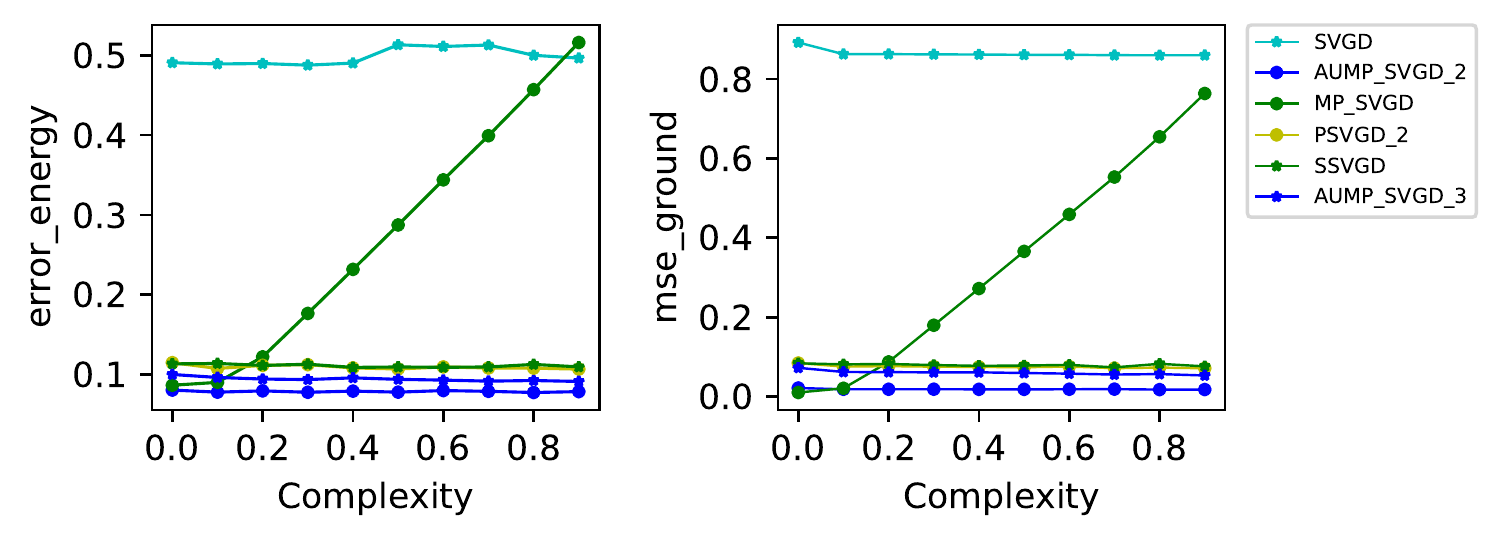}
    \caption{MSE and energy distance. Each graph structure is averaged over ten experiments.}
    \label{complex_vs}
\end{figure}

\textbf{Non-sparse experiment} We set the dimension of the target distribution to 50 and systematically transfer the sparse covariance matrix to a non-sparse one by augmenting the correlations between different dimensions along the main diagonal. The results are presented in Figure ~\ref{complex_vs}.

\vspace{-0.2cm}
As the density of the probability map increases, the discrepancy between MP-SVGD and the actual target distribution gradually magnifies. 
Eventually, MP-SVGD succumbs to the curse of dimensionality. This phenomenon arises due to the fact that the dispersion force of particles in MP-SVGD primarily depends on the size of the Markov blanket. 
With increasing density, MP-SVGD encounters the same challenges associated with high-dimensional scenarios as observed in SVGD. However, as illustrated in Figure~\ref{complex_vs}, regardless of the density of the target distribution's graph structure, AUMP-SVGD remains unaffected by variance collapse. This is attributed to the repulsion force of particles in AUMP-SVGD being influenced by the artificially chosen Markov blanket size, underscoring the superiority of our algorithm over MP-SVGD.

\subsection{Conditioned Diffusion Process}
\begin{figure}[h]
    \centering
    \includegraphics[width=0.5\textwidth]{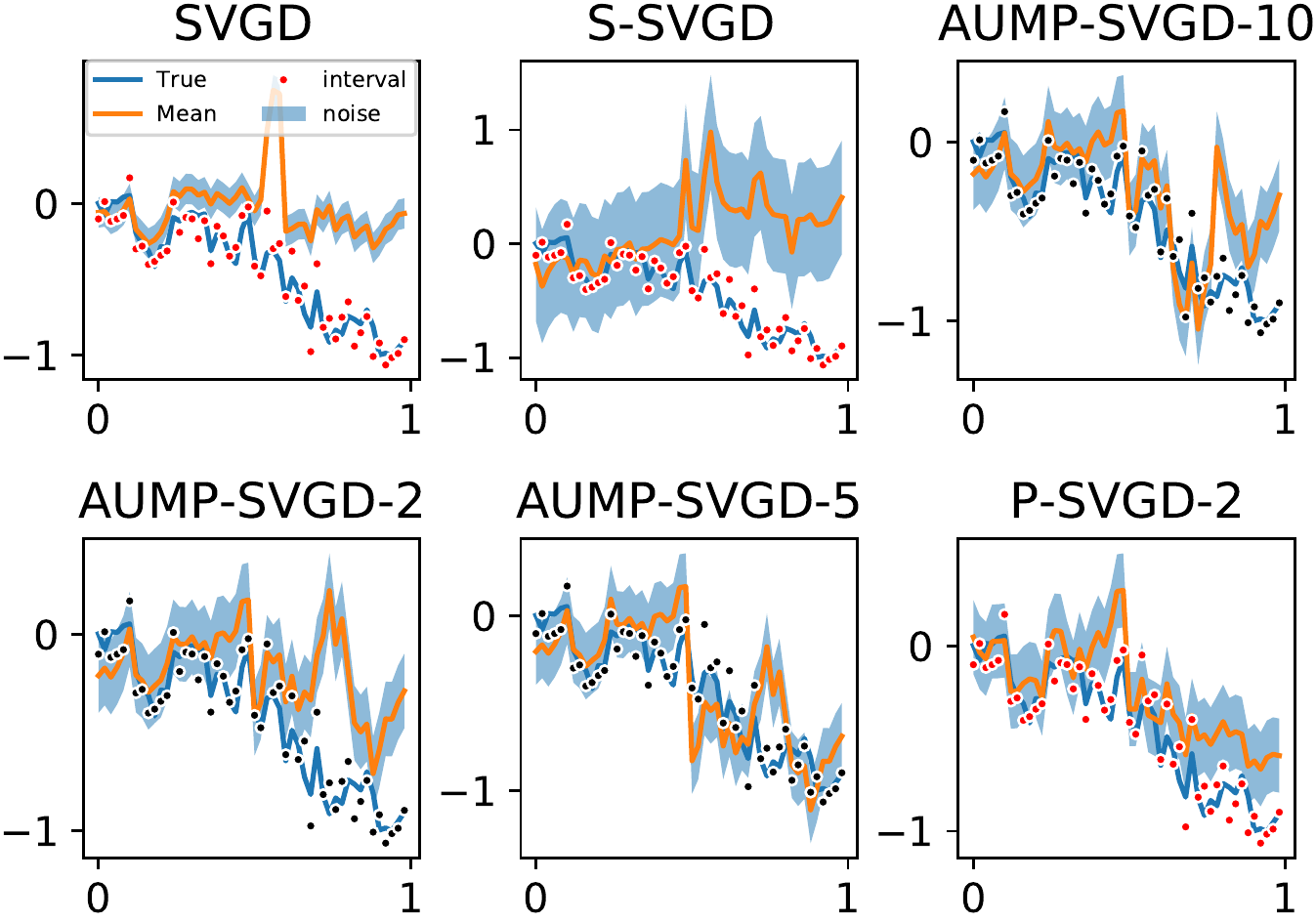}
    \caption{Results at iteration 1000.}
    \label{diff}
\end{figure}

The next example is a benchmark that is often used to test inference methods in high dimensions~(\citet{detommaso2018stein, chen2020projected, liu2022grassmann}). We consider a stochastic process $u:[0,1] \rightarrow \mathbb{R}$ governed by
\begin{equation*}
    \begin{small}
    d u=\frac{5 u\left(1-u^{2}\right)}{1+u^{2}} d t+d x, \quad u_{0}=0,
    \end{small}
\end{equation*}  
where $t \in(0,1]$, the forcing term  $x=\left(x_{t}\right)_{t \geq 0}$ follows a Brownian motion so that $x \sim \mathcal{N}(0, B)$ with  $B\left(t, t^{\prime}\right)=\min \left(t, t^{\prime}\right)$. The noisy data $\mathbf{z}=\left(z_{t_{1}}, \ldots, z_{t_{50}}\right)^{T} \in \mathbb{R}^{50}$ at 50 equi-spaced time points with $t_{i}=0.02 i$, where  $z_{t_{i}}=u_{t_{i}}+\epsilon$ for $\epsilon \sim \mathcal{N}\left(0, \sigma^{2}\right)$ with $\sigma=0.1$. The objective is to use $z$ to infer the forcing term $x$ and thus the state of the solution $u$. The results are given in Figure~\ref{diff} where the shadow interval depicts the mean plus/minus the standard deviation.

\vspace{-.2cm}

As depicted by Figure~\ref{diff}, it is evident that in the case of 50 dimensions, both SVGD and S-SVGD exhibit certain deviations from the ground truth while S-SVGD exhbits excessively large variances in numerous tests. Consequently, SVGD and S-SVGD prove inadequate in effectively addressing the conditional diffusion model. Conversely, our AUMP-SVGD demonstrates satisfactory performance with the size of $\Gamma_d$ of 5 and 10.

\subsection{Bayesian Logistic Regression}

We investigate a Bayesian logistic regression model from~\cite{liu2016stein} applied to the Covertype dataset from~\cite{asuncion2007uci}. 
We use 70\% data for training and 30\% for testing. We compare AUMP-SVGD with SVGD, S-SVGD, PSVGD, MP-SVGD, and Hamiltonian Monte Carlo(HMC) and the number of generated samples ranges from 100 to 500. Each experiment uses ten different random seeds and the error of each value does not exceed 0.01. We verify the impact of different sampling methods on the prediction accuracy and the results are given by Table~\ref{tab}. 

\begin{table}[h]
\centering
\caption{The Optimal values for each case are in bold.}\label{tab}
        {\scriptsize
    \begin{tabular}{l|lllllll}
    \hline
    {\scriptsize methods}    &  \multicolumn{7}{c}{\scriptsize accuracy}  \\ \hline
    \# {\scriptsize particles} & {\scriptsize HMC}  & {\scriptsize SVGD} & {\scriptsize S-SVGD} & {\scriptsize MP-SVGD} & {\scriptsize PSVGD\_2}   & {\scriptsize AUMP-SVGD-5} & {\scriptsize AUMP-SVGD-10}                                          \\ \hline
    100       & 0.70 & 0.74 & 0.76  & 0.74    & \textbf{0.77} & 0.73        & 0.75          \\
    200       & 0.71 & 0.74 & 0.762 & 0.74    & \textbf{0.79} & 0.78        & 0.78          \\
    300       & 0.73 & 0.74 & 0.762 & 0.741   & \textbf{0.81} & 0.80        & \textbf{0.81} \\
    400       & 0.80 & 0.75 & 0.76  & 0.74    & \textbf{0.83} & 0.81        & 0.81          \\
    500       & 0.81 & 0.75 & 0.765 & 0.75    & 0.85          & 0.82        & \textbf{0.86} \\ \hline
    \end{tabular}
        }
\end{table}

Table~\ref{tab} demonstrates that our AUMP-SVGD has a prediction accuracy rate similar to that of PSVGD, and as the number of particles increases, our algorithm is more accurate than other algorithms.

%% file: conclusion.tex
\section{Conclusion and Future work}
In this paper, we analyze the upper bound of the variance collapse of SVGD when the number of particles is finite. We show that the distribution of particles is restricted to a specific region rather than the entire probability space. We also propose the AUMP-SVGD algorithm to further overcome the dependency of MP-SVGD on the known and sparse graph structure. We show the effectiveness of AUMP-SVGD through various experiments. For future work, we aim to further investigate the convergence of SVGD with finite particles and tighten the estimation limit. We also plan to apply MP-SVGD to more complex real-world applications, such as posture estimation~(\cite{pacheco2014preserving}).

%% file: appendix.tex
\section*{Appendix A}
    \subsection*{Proof of Proposition 1 }
According to the fixed points assumption (A1),
    \[
    \Delta(\mathbf{x}_i) = \frac{1}{N} \sum_{j=1}^{N} \nabla_{\mathbf{x}_{j}} \log p\left(\mathbf{x}_{j}\right) k\left(\mathbf{x}_{i}, \mathbf{x}_{j}\right)+\nabla_{\mathbf{x}_{j}} k\left(\mathbf{x}_{i}, \mathbf{x}_{j}\right)= 0.
    \]
Then,
\begin{equation}
\begin{aligned}
    \mathbb{E}\left(\Delta(\mathbf{x}i)\right) & =\mathbb{E} (\nabla{\mathbf{x}} \log p\left(\mathbf{x}\right) k\left(\mathbf{x}_{i}, \mathbf{x}\right))+\nabla{\mathbf{x}} k\left(\mathbf{x}_{i}, \mathbf{x}\right) \\
    &= \mathbb{E}\left(\nabla{\mathbf{x}}\log p\left(\mathbf{x}\right)\right)\mathbb{E}\left( k\left(\mathbf{x}_{i}, \mathbf{x}\right)\right)+\mathbb{E}(\nabla{\mathbf{x}} k\left(\mathbf{x}_{i}, \mathbf{x}\right)) \\
    &= \mathbb{E}(\nabla{\mathbf{x}}k\left(\mathbf{x}_{i}, \mathbf{x}\right)) = 0. \label{pf1}
\end{aligned}
\end{equation}
In Equation~\eqref{pf1}, $\nabla{\mathbf{x}} \log p\left(\mathbf{x}\right)$ and $k\left(\mathbf{x}_{i}, \mathbf{x}\right)$ are independent and,
    \begin{equation*}
    \mathbb{E}\left(\nabla{\mathbf{x}} \log p\left(\mathbf{x}\right)\right) = \int \frac{p^{'}(\mathbf{x})}{p(\mathbf{x})} p(\mathbf{x}) dx \
    = \int p^{'}(\mathbf{x}) dx \
    = 0.
    \end{equation*}
    Associated with the Jensen gap (\citet{gao2017bounds}), we get
    \begin{equation*}
    \begin{aligned}
    | \mathbb{E}[e ^{\frac{-\left |\mathbf{x}_i - \mathbf{x} \right |_2^2 }{h} }\frac{2(\mathbf{x}_i-\mathbf{x})}{h}] - e^{\frac{-|\mathbf{x}_i - \mathbb{E}\mathbf{x}|_2^2}{h}} \frac{2(\mathbf{x}_i-\mathbb{E}\mathbf{x})}{h} | \
    \leq 2M\mathbb{E}\left| \mathbf{x} \right|,
    \end{aligned}
    \end{equation*}
    which in turn gives
\[
    \left | e ^{\frac{-\left \|\mathbf{x}_i - \mathbb{E}\mathbf{x} \right \|_2^2 }{h} }\frac{2(\mathbf{x}_i-\mathbb{E}\mathbf{x})}{h}  \right |< 2M\mathbb{E}\left | \mathbf{x} \right |
\]
since $\left|\mathbf{x}_i - \mathbb{E}\mathbf{x}\right|_2^2 = \mathrm{tr}\left((\mathbf{x}_i-\mathbb{E}\mathbf{x})(\mathbf{x}_i-\mathbb{E}\mathbf{x})^T\right) \leq c_0 \mathrm{tr}(\Sigma)$, where $c_0$ is a positive number.

\subsection*{Proof of Proposition 2 }

According to Proposition 1:
\[
\left \| \mathbf{x}  \right \|_2^2 \leq K^2 tr \Sigma
\]
Apply the expectation version of the Bernstein inequality~(\citet{banna2016bernstein}) for the sum of mean zero random matrices $\mathbf{x} _i\mathbf{x} _i^T-\Sigma$ and we obtain,
\begin{equation*}
\begin{small}
\begin{aligned}
\mathbb{E} \| \Sigma_m - \Sigma\| & = \frac{1}{m} \| \sum_{i=1}^{m} \mathbb{E} (\mathbf{x}_i\mathbf{x}_i^T - \Sigma) \| \\
& \leq 30 v \sqrt{n \log D}+4 M c^{-1 / 2} \sqrt{\log D} +M \gamma(c, n) \log D,
\end{aligned}
\end{small}
\end{equation*}
and $M$ is any number chosen such that,
\begin{equation*}
\begin{small}
\|\mathbf{x}\mathbf{x}^T - \Sigma \| \leq M. 
\end{small}
\end{equation*}
To bound $M$ is simple:
\begin{equation*}
\begin{small}
\|(\mathbf{x}\mathbf{x}^T - \Sigma) \|\le \| \mathbf{x} \|_2^2 + \|\Sigma \| \le 2K^2tr\Sigma,
\end{small}
\end{equation*}
which completes the proof.

\subsection*{Proof of Proposition 3}
Note that 
\begin{equation*}
\nabla_{\epsilon} \mathrm{KL}\left( q_{[T]}\left(\mathbf{x}_{\Gamma_{d}},\mathbf{x}_{\mathbf{S_{d}}}\right) \|p\left(\mathbf{x}_{\Gamma_{d}},\mathbf{x}_{\mathbf{S_{d}}}\right) \right) = \nabla_{\epsilon} \mathrm{KL}\left( q_{[T]}\left(\mathbf{x}_{\Gamma_{d}}\mid \mathbf{x}_{\mathbf{S_{d}}}\right) q\left(\mathbf{x}_{\mathbf{S_{d}}}\right) \| p\left(\mathbf{x}_{\Gamma_{d}} \mid \mathbf{x}_{\mathbf{S_{d}}}\right) p\left(\mathbf{x}_{\mathbf{S_{d}}}\right) \right).
\end{equation*}
Now we derive the optimality condition for Equation~\eqref{argmin}. Note that,
\begin{equation*}
\mathrm{KL}\left( q_{[T]}\left(\mathbf{x}_{\Gamma_{d}},x_{\mathbf{S_{d}}}\right) \|p\left(\mathbf{x}_{\Gamma_{d}},x_{\mathbf{S_{d}}}\right) \right) =\mathbb{E}_{q\left(\mathbf{x}_{\neg d}\right)}\left[ \mathrm{KL}\left( q_{[T]}\left(x_{\Gamma_d} \mid \mathbf{x}_{\neg d}\right) \|p\left(z_{\Gamma d} \mid \mathbf{x}_{\neg d}\right) \right) \right].
\end{equation*}

Following the proof of Theorem 3.1 by~\citet{liu2016stein}, we have
\begin{equation*}
\begin{split}
\nabla_{\epsilon} \mathrm{KL}\left( q_{[T]}\left(\mathbf{x}_{\Gamma_{d}},\mathbf{x}_{\mathbf{S_{d}}}\right) \|p\left(\mathbf{x}_{\Gamma_{d}},\mathbf{x}_{\mathbf{S_{d}}}\right) \right) |_{\epsilon=0} & =  \\
& -\mathbb{E}_{q\left(\mathbf{y}_{\Gamma_{d}} \mid \mathbf{y}_{\neg d}\right)}\left[ \phi_{\Gamma_{d}}\left(\mathbf{y}_{\neg d}\right) \nabla_{\mathbf{y}_{\Gamma_{d}}} \log p\left(\mathbf{y}_{\Gamma_{d}} \mid \mathbf{y}_{\neg d}\right)+\nabla_{\mathbf{y}_{\Gamma_{d}}} \phi_{\Gamma_{d}}\left(\mathbf{y}_{\neg d}\right) \right].
\end{split}
\end{equation*}
According to~\cite{liu2016stein}, we can show that the optimal solution is given by $\frac{\phi_{\Gamma_d}^{*}}{\|\phi_{\Gamma_d}^{*}\|_{\mathcal{H}_{\Gamma_d}}}$, where
\begin{equation*}
\begin{aligned}
\phi_{\Gamma_{d}}^{*}\left(\mathbf{x}_{\neg d}\right) = \mathbb{E}_{\mathbf{y}_{\neg d} \sim q}\left[ k_{\Gamma_{d}}\left(\mathbf{x}_{\neg d}, \mathbf{y}_{\neg d}\right) \nabla_{\mathbf{y}_{\Gamma_{d}}} \log p\left(\mathbf{y}_{\Gamma_{d}} \mid \mathbf{y}_{\neg d}\right) + \nabla_{\mathbf{y}_{\Gamma_{d}}} k_{\Gamma_{d}}\left(\mathbf{x}_{\neg d}, \mathbf{y}_{\neg d}\right) \right].
\end{aligned}
\end{equation*}

\subsection*{Proof of Proposition 4}
Similar to the last proof, first we have
\begin{equation*}
\begin{small}
\nabla_{\epsilon}  \mathrm{KL} \left[\frac{q_{[T]}(x_{d} \mid \mathbf{x}_{\mathbf{S_{d}}} )} {q(\mathbf{x}_{\neg d})}  \| \frac{p(x_{d} \mid \mathbf{x}_{\mathbf{S_{d}}})} {p(\mathbf{x}_{\neg d})} \right] = \nabla_{\epsilon} \mathrm{KL}\left(q_{[T]}\left(x_{d}\mid \mathbf{x}_{\mathbf{S_{d}}}\right) q\left(\mathbf{x}_{\mathbf{S_{d}}}\right) \| p\left(x_{d} \mid \mathbf{x}_{\mathbf{S_{d}}}\right) q\left(\mathbf{x}_{\mathbf{S_{d}}}\right)\right).
\end{small}
\end{equation*}
Then we derive the optimality condition $\phi_{d}^{*}$ for Equation~\eqref{p2argmin}, 
\begin{equation*}
\begin{small}
\mathbf{KL}\left(q_{\left[T\right]}\left(x_{d},\mathbf{x}_{\mathbf{S_{d}}}\right) \|p\left(x_{d},\mathbf{x}_{\mathbf{S_{d}}}\right)\right) =\mathbb{E}_{q\left(\mathbf{x}_{S_{d}}\right)}\left[\operatorname{KL}\left(q_{\left[T\right]}\left(x_{d} \mid \mathbf{x}_{S_{d}}\right) \| p\left(x_{d} \mid \mathbf{x}_{S_{d}}\right)\right)\right] 
\end{small}
\end{equation*}
Following the proof of Theorem 3.1 in~\cite{liu2016stein}, we have
\begin{equation*}
\begin{small}
\nabla_{\epsilon} \mathrm{KL}\left(q\left(x_{d},\mathbf{x}_{\mathbf{S_{d}}}\right) \|p\left(x_{d},\mathbf{x}_{\mathbf{S_{d}}}\right)\right) |_{\epsilon=0} = -\mathbb{E}_{q\left(y_{d} \mid \mathbf{y}_{S_{d}}\right)}\left[\phi_{d}\left(\mathbf{y}_{C_{d}}\right) \nabla_{y_{d}} \log p\left(y_{d} \mid \mathbf{y}_{S_{d}}\right)+\nabla_{y_{d}} \phi_{d}\left(\mathbf{y}_{S_{d}}\right)\right].
\end{small}
\end{equation*}

According to~\cite{liu2016stein}, we can show that the optimal solution is given by $\frac{\phi_{d}^{*}}{\left\|\phi_{d}^{*}\right\|_{\mathcal{H}_{d}}}$, where
\begin{equation*}
\begin{aligned}
\phi_{d}^{*}\left(\mathbf{x}_{C_d}\right) = \mathbb{E}_{\mathbf{y}_{C_d} \sim q}\left[ k_{S_{d}}\left(\mathbf{x}_{C_d}, \mathbf{y}_{C_d}\right) \nabla_{\mathbf{y}_{S_{d}}} \log p\left(\mathbf{y}_{S_{d}} \mid \mathbf{y}_{C_d}\right) + \nabla_{\mathbf{y}_{S_{d}}} k_{S_{d}}\left(\mathbf{x}_{C_d}, \mathbf{y}_{C_d}\right) \right],
\end{aligned}
\end{equation*}
which completes the proof.

\section*{Appendix B}

\subsection*{Numerical verification of $m$-dependent}

In our paper, we use the concept of $m$-dependent in the mixture to estimate the variance of non-i.i.d. particles. Here we give a verification experiment for $m$-dependent.

\begin{figure}[h]
    \centering
    \includegraphics[width=0.5\textwidth]{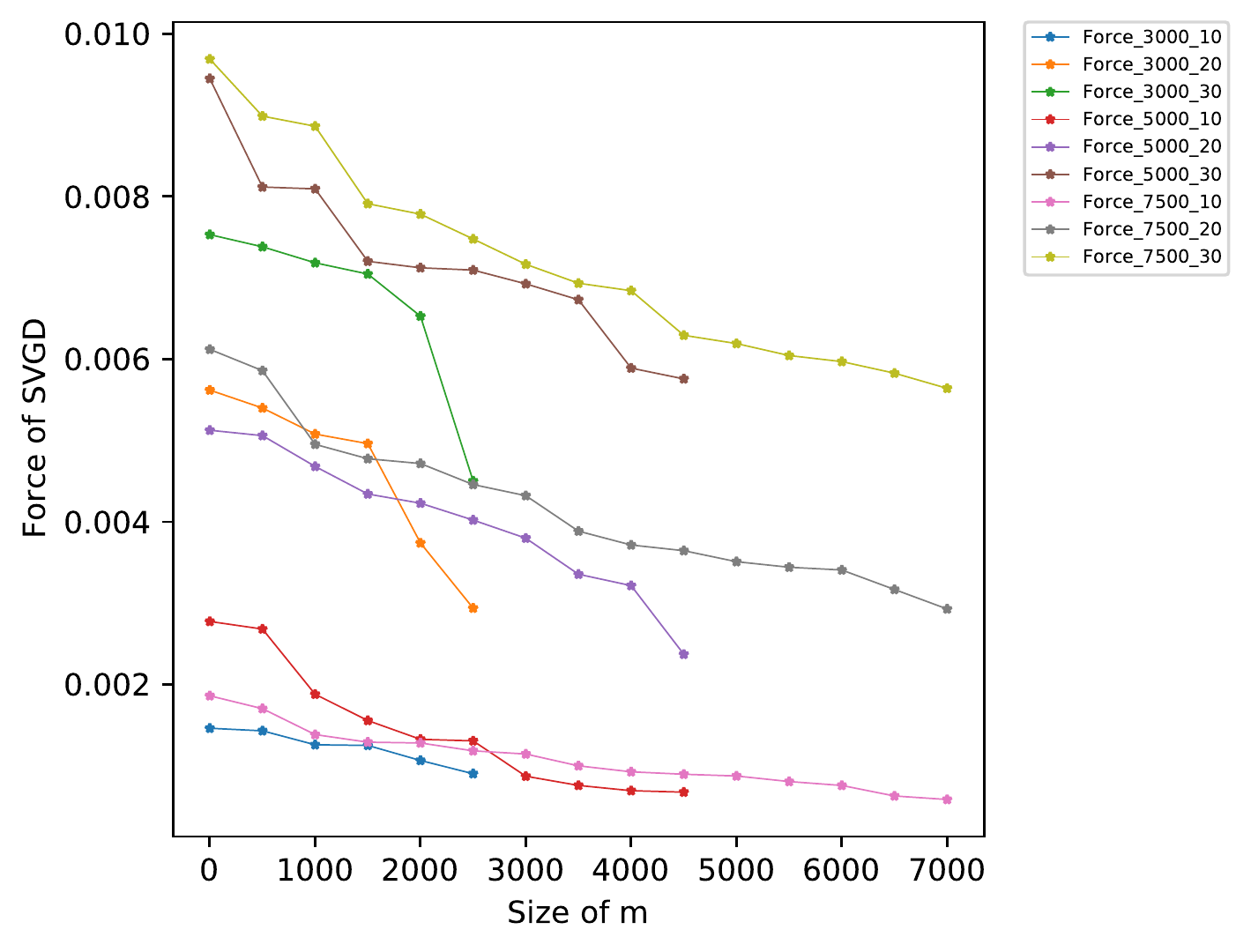}
    \label{m-de}
\end{figure}

The above figure shows the final dynamic magnitude of the two particles in SVGD. Here, the legend $\text{SVGD}\_3000\_10$ indicates 3000 particles for a 10-dimensional Gaussian model. We sort the particles according to $\|\mathbf{x}\|_2$ and measure the update effect between two particles at intervals of 500, i.e.,

\begin{equation}
\begin{small}
\frac{1}{N}  \nabla_{\mathbf{x}_{n}^{\ell}} \log p\left(\mathbf{x}_{n}^{\ell}\right) k\left(\mathbf{x}_{n}^{\ell}, \mathbf{x}_{m}^{\ell}\right)+\nabla_{\mathbf{x}_{n}^{\ell}} k\left(\mathbf{x}_{n}^{\ell}, \mathbf{x}_{m}^{\ell}\right). 
\end{small}
\end{equation}

We can see from the above figure, as the number of particle intervals increases, the force between particles decreases.

\subsection*{Numerical verification of Proposition 1-2}

In the presented tabular data, we have undertaken empirical investigations for the target distribution $\mathbf{N}\left(0, I_{d i m}\right)$ and have demonstrated the soundness of our theoretical bounds through these experiments.
\begin{table}[H]
\centering
    \caption{Upper bound of  $\|x\|_{2}^{2}$  in Proposition 1}
    \begin{tabular}{ll|llllll}
    \hline
    \multicolumn{2}{l|}{dim}                                                 & 2    & 5    & 10   & 15   & 20    & 25   \\ \hline
    \multicolumn{1}{l|}{\multirow{2}{*}{10 particles}} & max $\|x\|_{2}^{2}$ & 1.43 & 1.18 & 1.17 & 1.17 & 1.17  & 1.17 \\
    \multicolumn{1}{l|}{}                              & theoretical bound         & 4.82 & 5.56 & 5.39 & 5.36 & 5.35  & 5.35 \\ \hline
    \multicolumn{1}{l|}{\multirow{2}{*}{50 particles}} & Max $\|x\|_{2}^{2}$ & 2.11 & 1.63 & 1.55 & 1.52 & 1.52  & 1.51 \\
    \multicolumn{1}{l|}{}                              & theoretical bound         & 9.41 & 16.1 & 17.1 & 14.8 & 14.31 & 14.0 \\ \hline
    \end{tabular}
\end{table}

In the presented table, the upper bound of the variance error of SVGD has been assessed for the case where the distribution is  $\mathbf{N}\left(0, I_{d i m}\right)$.

\begin{table}[H]
\centering
    \caption{Upper bound of  $\mathbb{E}\left\|\Sigma_{m}-\Sigma\right\|_{2}^{2}$  in Proposition 2
    }
    \begin{tabular}{ll|lllll}
    \hline
    \multicolumn{2}{c|}{Number of particles}                                                  & 1000  & 5000  & 10000 & 15000  & 20000 \\ \cline{1-2}
    \multicolumn{1}{l|}{\multirow{2}{*}{Dim-2}} & $\mathbb{E}\|\Sigma_{m} - \Sigma\|_{2}^{2}$ & 0.008 & 0.002 & 0.004 & 0.0063 & 0.003 \\
    \multicolumn{1}{l|}{}                       & theoretical bound                                 & 0.082 & 0.019 & 0.01  & 0.0078 & 0.005 \\ \hline
    \multicolumn{1}{l|}{\multirow{2}{*}{Dim-5}} & $\mathbb{E}\|\Sigma_{m} - \Sigma\|_{2}^{2}$ & 0.3   & 0.11  & 0.06  & 0.06   & 0.04  \\
    \multicolumn{1}{l|}{}                       & theoretical bound                                 & 4.63  & 1.71  & 1.02  & 0.71   & 0.57  \\ \hline
    \end{tabular}
\end{table}